# Latent patterns of urban mixing in mobility analysis across five global cities

Zhuangyuan Fan[1], Becky P.Y. Loo[1,2*], Fabio Duarte[3], Carlo Ratti[3,4], and Esteban Moro[5,6]

[1] Department of Geography, The University of Hong Kong, Hong Kong SAR, China
[2] Institute of Marine Sustainable Development, Liaoning Normal University, Dalian, China
[3] Senseable City Lab, Massachusetts Institute of Technology, Cambridge, MA, USA
[4] ABC Department, Politecnico di Milano, Milan, Italy
[5] Network Science Institute and Department of Physics, Northeastern University, Boston, MA, USA
[6] Media Lab, Massachusetts Institute of Technology, Cambridge, MA, USA
* Corresponding: bpyloo@hku.hk

## Abstract

This study leverages large-scale travel surveys for over 200,000 residents across Boston, Chicago, Hong Kong, London, and São Paulo. With rich individual-level data, we make systematic comparisons and reveal patterns in social mixing, which cannot be identified by analyzing high-resolution mobility data alone. Using the same set of data, inferring socioeconomic status from residential neighborhoods yield social mixing levels 16% lower than using self-reported survey data. Besides, individuals over the age of 66 experience greater social mixing than those in late working life (aged 55–65), lending data-driven support to the "second youth" hypothesis. Teenagers and women with caregiving responsibilities exhibit lower social mixing levels. Across the five cities, proximity to major transit stations reduces the influence of individual socioeconomic status on social mixing. Finally, we construct detailed spatio-temporal place networks for each city using a graph neural network. Inputs of home-space, activity-space and demographic attributes are embedded and fed into a supervised autoencoder to predict individual exposure vectors. Results show that the structure of individual activity space, i.e., where people travel to, explains most of the variations in place exposure, suggesting that mobility shapes experienced social mixing more than sociodemographic characteristics, home environment, and transit proximity. The ablation tests further discover that, while different income groups may experience similar levels of social mixing, their activity spaces remain stratified by income, resulting in structurally different social mixing experiences.

## Main

Human mobility plays a central role in many of the 21st century's urban challenges[1–3], including carbon emissions[4,5], air pollution[6], non-renewable energy consumption[7], traffic congestion[8,9], noise pollution[10,11], road fatalities and casualties[12], transport inequity[7], and disease spread[13].

Yet, mobility also shapes other important dimensions of urban life, such as experienced social mixing—the diversity of people an individual encounters as one moves through the city. From commutes to grocery trips, daily mobility creates opportunities for contact across socioeconomic lines. Whether and how these paths intersect matter: a broad body of research links experienced social mixing to creativity[14], health outcomes[15], job opportunity and economic mobility[16,17]. The United Nations Human Settlements Program's (UN-Habitat) Sustainable Development Goal (SDG) 11[18] specifically promotes the idea of inclusive human settlements to ensure that cities are equitable for all residents. Nevertheless, persistent patterns of social segregation remain in cities[19,20], and the factors that shape our everyday social exposure continue to be debated in research and in practice[3,20,21].

Over the past decade, the rapid growth of urban data has transformed research on segregation and mixing, especially in cities across the United States[3,19]. However, most studies have relied on high-resolution mobility traces (such as those from mobile phones and smart cards) while inferring socioeconomic status (notably income and age) primarily from estimated home neighborhoods or frequently visited places (e.g., rental apartments, schools, and senior housing. A detailed literature review is included in SI Table S3)[19,22,23]. Such home-neighborhood or visit-based strategies of socioeconomic proxies are approximate and much less applicable in compact cities such as Hong Kong and Singapore, where people with different income levels and/or ethnicities often live in close proximity[20,24,25]. Moreover, the high density and diversity of land use also makes using mobile phone traces to identify school visits (often used to infer age and/or child caring responsibilities) or type of housing (e.g. affordable housing to infer income) challenging beyond broad categories of home, work and other activity locations[26].

This study takes a different approach by leveraging large-scale travel surveys that include detailed individual-level socioeconomic information. It is recognized that official territory-wide travel surveys also suffer from issues like non-response and underreporting[27-30]. Yet, the rich and detailed socioeconomic and trip data allow researchers to identify latent patterns in social mixing which can only be revealed by disaggregated individual-level data. Hence, rather than maximizing spatial and temporal detail from mobility traces, we draw on the granular socioeconomic and trip data provided by travel surveys, which capture rich demographic and behavioral information consistently across cities. By analyzing more than 200,000 travel diary entries, we provide the first cross-continental study of how daily mobility shapes social mixing across five contrasting global cities—Boston, Chicago, Hong Kong, London, and São Paulo.

# Results

The five travel survey data included are the Massachusetts Travel Survey (Boston), My Daily Travel (Chicago), Travel Characteristics Survey (Hong Kong), London Travel Diary (LTD) and São Paulo Travel Diary. These surveys were published between 2011 to 2018. The coverage, resolution, and survey time period can be found from Table 2. The number of participants interviewed for each survey ranges from 26,770 to 101,385. When compared to the size of the local population as of the year of each survey, the coverage ranged from 0.29% to 0.53%. As they were official surveys conducted by professional survey teams, expansion factors have been provided to correct sampling bias and for data expansion.

## Measure social mixing using travel survey data

The first goal of the research is to reveal the impact on levels of social mixing using the individual self-reported household income in travel surveys in comparison to home-location-inferred income. Our primary measure, the daytime mixing ($DM_i$), is adapted from Moro et al.[3] and quantifies how evenly an individual encounters people from different income groups during the day (excluding trips back home) (Fig. 1a). To calculate $DM_i$, we first rank participants within each city by their reported household income and then assign them to one of four groups: low, medium-low, medium-high, or high. The four income groups within each city have relatively similar proportions after applying the expansion factors to all participants (See SI for details). Next, using the recorded stops from travel surveys, we identify when participants might have encountered others during the day. Because each city defined travel stops differently (e.g., by street blocks, coordinates, or census blocks), we standardize the location unit by using the H3 hexagon index[31] as our primary definition of place. With an individual's income group $i$ and a series of places that they visited, we can compute $DM_i$ using Equation (1): $DM_i = \frac{2}{3}\sum_q |\tau_{iq} - \frac{1}{4}|$, where $\tau_{iq}$ is individual $i$'s relative exposure to income group $q$. Here $\tau_{iq}$ is defined as Equation (2): $\tau_{iq} = \sum_\alpha \tau_{i\alpha} \tau_{\alpha q}$, where $\tau_{i\alpha}$ is individual $i$'s proportion of visits at a place $\alpha$ among all places that $i$ visited. $\tau_{\alpha q}$ is all people from the income group $q$'s proportion of total visits at place $\alpha$. A $DM_i = 1$ means that an individual meets other income groups evenly throughout the day. While a $DM_i = 0$ indicates that an individual only meets people of the same income group in places he/she visited in a day. We demonstrate the robustness of our primary measure of $DM$ against the definitions of trip stops, income groups and weighting factors in SI Section 2. Further, we show that the main findings in this study are not sensitive to the definition of collocation, including taking into consideration of the estimated time that people will meet each other, and the spatial unit of place (See 22 variations of social mixing measure for robustness test in SI Table S2). It is worth noting that people together at a place does not necessarily imply social interaction, though physical collocation is a prerequisite for any in-person interaction to take place. Previous studies suggest that people staying in the same place, especially for a longer duration, are more likely to interact (e.g., having conversations) and engage in group activities[16,32-33]. In this paper, we describe the collocation as social mixing and recognize that social interaction is only a potential outcome.

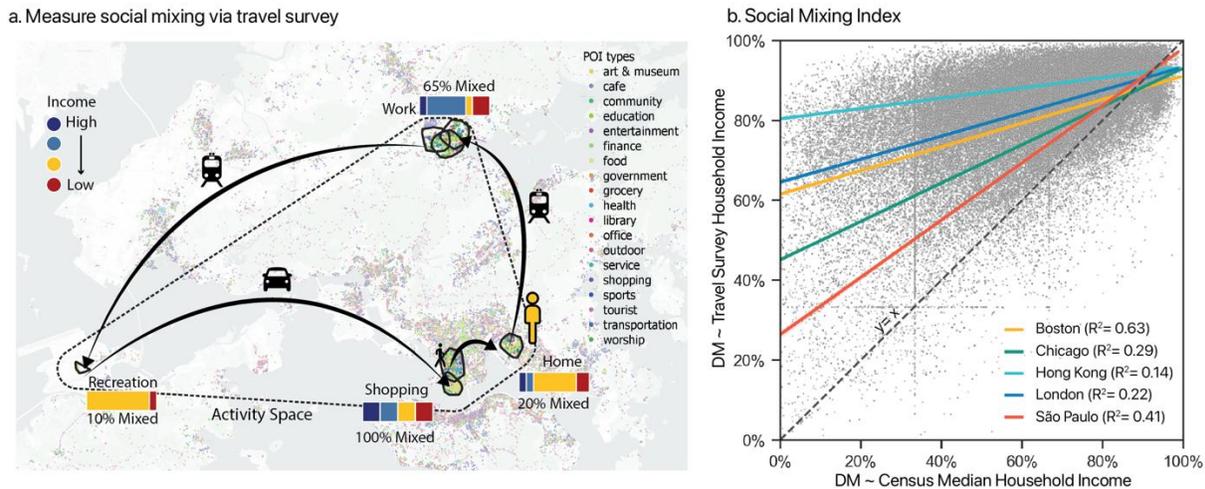

*Fig. 1 Measures of the daytime and nighttime social mixing for five cities. a) Measure the daytime social mixing. An individual visits different places and encounter others from a different income group, thus experience different level of social mixing. b) Compare the daytime social mixing (DM) measured using census survey median household income and reported income level in the travel survey. The dash line indicates y = x. For each city, we fit a simple ordinary least square regression line to show the relationship between y and x. The scatter plot only includes participants over age 11 and travelled at least once in the survey day (Boston n = 7,626, Chicago n = 18,258, Hong Kong n = 33,289, London n = 30,381, Sao Paulo n = 46,957).*

Next, as a comparison, we extract the median household income (from local census surveys) based on each participant's home location. With the approximated household income, we re-estimate the associated *DM* index by replacing the participant's income group with the home-location-inferred income group (see Methods for details on the census data source and city-specific resolution). Fig. 1b shows that when we replaced the survey income with median household income from respondents' neighborhoods, *DM* fell by an average of 16%. In Chicago—a city known for its residential segregation (with its night-time social mixing (NM) level around 58%)—the survey-based *DM* is 12% higher than the census tract-based estimate. This result shows that, using the same set of survey data, home neighborhood proxies tend to estimate lower social mixing levels than using self-reported data.

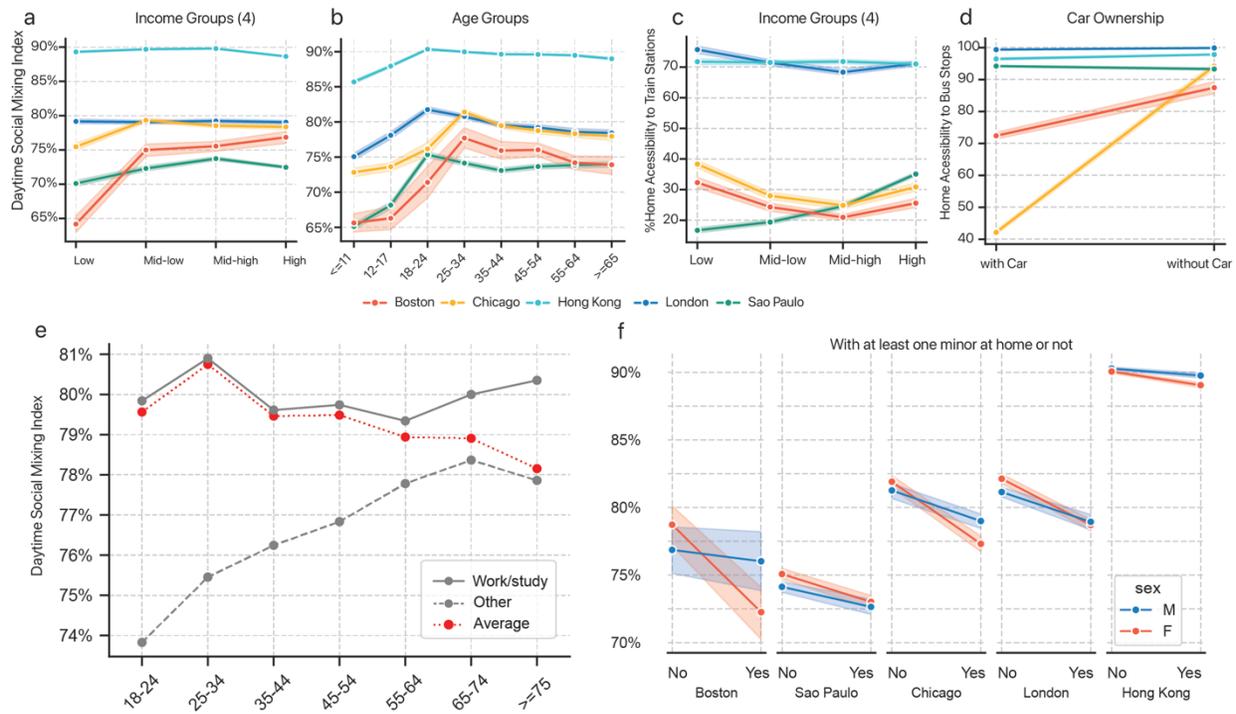

Fig. 2 Explain daytime social mixing. *a. DM is lower for low-income groups than for mid-income groups in Boston (n= 5,894), Chicago (n = 15,738), and São Paulo (n = 40,896), while gaps are less pronounced in Hong Kong (n = 51,619) and London (n = 30,277). Only people with age>17 are included. b. DM distribution by age and city. c. Home accessibility to train stations by income group in each city. d. Home accessibility to bus stops by household car ownership. e. Average DM across all cities by age groups and work status. DM declines gradually with age (dashed line), but rises again after retirement (65–74). Working individuals report higher DM than non-working peers, especially at younger ages. f. DM distribution by gender and care-taking responsibilities at early career age. Only person between age 21 to 45 are included (Boston n = 2,064, Chicago n = 8,365, Hong Kong, n = 25,594, London, n = 15,394, São Paulo, n = 20,227). In a-d, f, the solid dot marker shows the mean value of the sample by each group. The error bar area represents the 95% confidence interval. Overall income distributions with different definitions are shown in Fig. S3-4. Details of work status standardization across different surveys are included in Table S9.*

## Patterns of social mixing across the life course: age, gender, and caregiving responsibilities

Besides income, travel surveys also provide other rich socioeconomic attributes. Previous literature has shown that vulnerable population, such as teenagers and low-income groups, tend to be more socially segregated[19-21,23]. As age is generally missing from the mobile-phone based records, these studies have used proxies such as school visits to estimate people's age[23]. Focusing on the vulnerable population, we check the *DM* distribution across income groups, age groups and caregiving responsibilities (by gender) leveraging the self-reported income, age and gender information in travel surveys. Fig. 2a shows that people in the low-income group are likely to be less socially mixed than mid-income groups in Boston, Chicago, and São Paulo, while such gaps among income groups are less pronounced in Hong Kong and London. Fig.

2b shows that teenagers (aged 12-17) exhibit significantly lower social mixing levels than young adults (aged 18-24). In Chicago and Boston, the highest social mixing occurs among individuals aged 25-34, whereas it peaks earlier among those aged 18-24 in the other cities.

Moreover, we found data support for the "second youth" or "third age" effect when combining the working status and age information. Even though $DM$ tends to decline gradually across all cities with age (red dash line in Fig. 2e), separating the actively working or studying population (full-time, part-time, student, or home making) from their counterparts reveals a reversal. Both groups show higher $DM$ immediately after the retirement age (group 65-74) compared with the late working age (55-64) (0.75% and 0.83% increase post-retirement age, respectively). While working/studying individuals on average experience more social mixing than their counterparts, such gap is much more significant at younger age. We repeat the analysis by testing each city separately and using an alternative measure of social mixing. The results are robust that $DM$ largely remains similarly high or increases immediately after the retirement age (group 65-74) in all five cities (See Fig. S8).

Another advantage of travel survey data is that we can combine information of age, household composition, and gender to identify people who are likely to be caretakers (often parents) and have caregiving responsibilities (see Methods). While men and women (per reported gender) are equally socially mixed in our selected cities, Fig. 2f implies that women with caregiving responsibilities experience a higher reduction in social mixing (ranging from 1.0-6.5% reduction) than their male counterparts (ranging from 0.5-2.3% reduction) (See a full statistical comparison in SI Table S5). Given that women with caregiving responsibilities are likely to be at an early career stage, this pattern suggests that unequal mobility constraints may be a contributing factor to broader gender disparities in career advancements (often described as the "leaky pipeline"), though further research is needed to disentangle this relationship from other structural barriers.

## Explaining social mixing: demographics versus place exposure

Moving beyond describing patterns, we look at how individuals' social background, mobility pattern and the associated urban structure influence their social mixing levels across the five cities using travel survey and other location data. We build a series of regression models that estimate social mixing based on five groups of variables (see Methods):

- Sociodemographic background (S): income, age, gender, caregiving responsibilities, employment status, vehicle ownership, and home public transit accessibility.

- Home-transit hub distance (H): the network distance of an individual's home zone to the nearest transit hubs with the highest inflow during morning peak hours. To avoid biases due to urban size, we allow for both mono-centric and poly-centric city concepts for transit hub identification following methods developed by Roth et al.[34]

- Place exposure at all destinations (PD): the aggregated number of POIs within a *t*-minute walk of all destinations of a person's daily trip, categorized into *N* place categories (e.g., restaurants, schools, retail, hospitals, etc.).

- Place exposure at the home zone (PH): mirroring PD, it measures the number of POIs grouped by categories within a *t*-minute walking time of an individual's home zone.

- Mobility pattern (M): percentage of trips taken by different transportation modes (public transit or private cars).

The full regression model is specified as Equation (3): $DM = \{S\} + \{H\} + \{PD\} + \{PH\} + \{M\}$. The unit of measurement is at individual level. Using the LMG method[35] (See SI Section 4 for details), we compute the relative importance of each variable group in Equation (3). Fig. 3a illustrates that place exposure at destinations explains 50% to 69% of the model predictive power across all five cities. Sociodemographic factors, place exposure at home locations, and distance from transit hubs generally have lower explanatory power for social mixing levels. Overall, individual sociodemographic factors contribute more in Chicago (12.5%) and Boston (8.9%) than the three other cities (<5%). Full variable coefficients are included in Fig. S15 and 16.

As the five cities in our study have different urban forms, we repeat the analysis by stratifying residents by distance to the nearest transit hub. Fig. 3b unveils an underlying pattern: for those living farther from major transportation hubs, sociodemographic factors are more important and can explain up to 76% of social mixing. For people living within 20 km of a major transit hub, their social mixing level is largely determined by place exposure at destinations (50-79%).

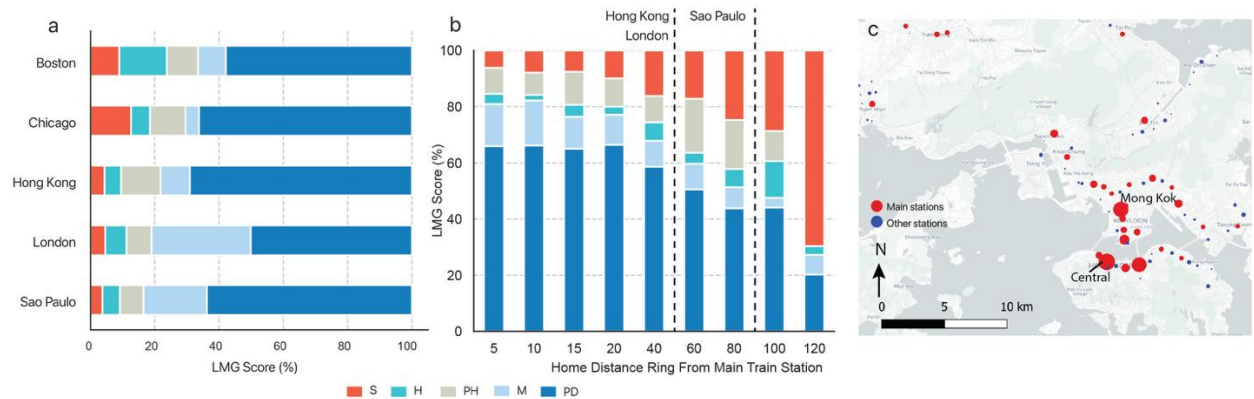

*Fig. 3. Explain DM. a. LMG score of Equation 3. Full regression results are included in the SI Table S6 and Fig. S16 and 17. We show a variation of this figure by limiting the samples only to people who live within 30 km of the downtown area in SI Fig. S15. In addition, Fig. S18 shows the results when normalizing the PD vector by average total POIs an individual visited. b. Average LMG score of the five cities when grouping people based on their home distance from their closest main transit stations. Dash lines indicate the total size of the specific city. Results by each city is shown in Fig. S15. c. Distribution of main train stations and other stations in Hong Kong. The size of the circle shows the weighted inflow of each station in the morning between 6 and 11 am.*

## Transportation modes

When summarizing the modal distribution of trips, we recognize that transit accessibility varies within the five cities across different income groups. Fig. 2c shows that accessibility to train stations is rather uniform across all income groups in Hong Kong and London. On the contrary, both the lowest and highest income groups in Chicago and Boston have higher access to train stations. In São Paulo, people with higher income have better access to train stations. Fig. 2d further describes the connection between vehicle ownership and access to bus stations. In Chicago and Boston, people who have at least one car at home are less likely to live closer to bus stations, while for Hong Kong, São Paulo, and London, home accessibility to bus stations is rather independent from household car ownership.

After controlling for related variables, we see that the percentage of trips taken by public transit is positively correlated with *DM* in all five cities. Specifically, a one percent increase of public transit trips is associated with 0.1 to 0.3% increase of *DM* (Table 1). In contrast, private car usage shows smaller or even negative associations with *DM*, indicating that private cars contribute less to fostering social mixing. Comparing Model 1 and Model 2 reveals that the positive effect of public transit weakens after accounting for place exposure variables (*PD* and *PH*). This suggests that the positive relationship between public transit and mixing in most cities can be partially explained by the types of places that transit users access—namely, places with higher levels of POI density and diversity. In London and Hong Kong, the weaker decline in coefficients implies that the link between public transit and *DM* operates more independently of home and destination characteristics, reflecting these cities' relatively high POI density across the broader urban fabric.

## Stratified activity spaces across income groups

Although our earlier results show that place exposure at destinations (*PD*) is the strongest predictor of individual-level social mixing, a remaining issue is that exposure itself is a result of the complex interplay of a city's underlying built environment, notably the distribution of POIs and transport network, and people's characteristics and mobility patterns. To tease out the different effects, we develop a machine learning framework to predict individual daily place exposure—represented as an *N*-dimensional vector of potential visits to urban amenities—using urban (transport) structure, home and activity spaces, and socioeconomic attributes.

Specifically, we construct a spatio-temporal place networks for each city (over 100,000 POIs and 12.7 million edges) and embed them using a graph neural network (GNN) to capture the full spatial context that a person may experience. Compared to using raw POI categories within the same space, GNN embeddings capture latent spatial context (e.g., street network and POI relationships into a dense vector). These embeddings, combined with demographics and travel features, are then fed into a supervised autoencoder designed to predict individual exposure vectors (Fig. 4a-c). Using this model, we evaluate three inputs systematically: home-space features ($h_h$), activity-space features ($h_a$), and demographic attributes ($h_d$). Home-space features represent the POIs accessible within a 15-minute walking distance from an individual's home. Activity-space features represent amenities within the convex hull encompassing all visited locations in a day. Both features were constructed using the representation of each POI aggregated from the GNN embeddings (See Method for identifying the adjacent POIs, definition of activity space, embedding aggregation, and model details). Demographic features capture socioeconomic information such as age, gender, and employment status. These inputs are processed separately through corresponding encoders, producing three embeddings that are concatenated into the comprehensive latent vector $z$ in a permutation fashion, summarizing essential demographic, home, and activity-space details.

To predict place exposure vector $Y$, we have the decoder as Equation (4): $\widehat{Y^{(i)}} = W_d z^{(i)} + b_d \quad \in R^N$, where $b_d$ is the bias term, $W_d$ is the weight matrix. To analyze the contribution of the three approaches of embedding, we conduct an ablation study by setting one of the $h_h$, $h_a$ or $h_d$ from the Equation (4) to zero and evaluate the change in reconstruction loss. We repeat the analysis by randomly reconstructing the train/test datasets 10 times and report the average results here (each time we use 75% data as training and 25% data as test). Each train and test data have equal distribution of people from different income groups.

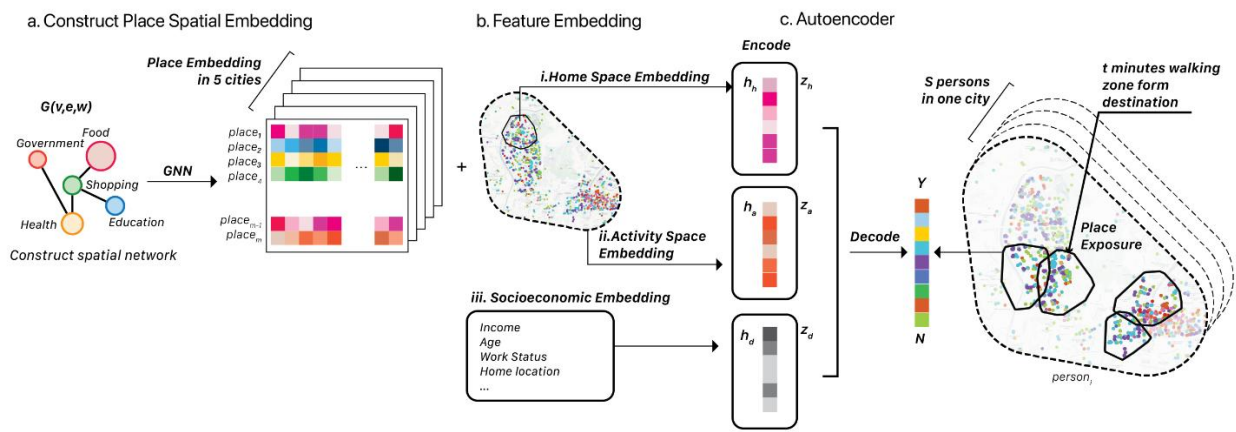

*Fig. 4 Conceptual diagram of an encoder-decoder architecture that predicts individual place exposure. a. Construct a POI place spatial network using distance and travel-time based method (See Method). Then use a GNN model, we encode all POI places in each city into an embedding vector. For each city, we have M POIs and the embedding dimension is 32. b. Create the feature embedding. For each person's home location, we construct a 15-minute walking isochrone and gets all available POI places within the isochrone. The home-space embedding ($h_h$) is the mean embedding of all POI places associated with their home location 15-minute walking isochrone. Similarly, for each person, we got their maximum travel activity space, a polygon that contains all locations that the person ever visited during the day. The activity-space embedding ($h_a$) is the mean embedding of all POI places within the person's activity space. Sociodemographic embedding includes all personal attributes such as age, gender, and work status. c. The autoencoder model takes the three embedding in a permutation fashion and compress them into a latent code z, and then decode to an output layer, which is the N-Dimensional POI place exposure vector (N is the total POI types).*

Fig. 5a visualizes the change of test data $mean - R^2$ for the three models in the five global cities. Here the $mean - R^2$ represents the average $R^2$ of each POI category in the place exposure vector (See SI Fig. 22 for model performance results measured by MAE and MSE). Results show that activity-space embeddings or mobility patterns are far more predictive than home-space or demographic features. Across cities, $h_a$ predicts about 60-80% (average of 59%) variance of the place exposure, compared to 25-45% for $h_h$ and <10% for $h_d$. Adding demographics provides marginal improvments. Robustness tests in SI Section 3 using alternative POI datasets and feature construction methods show consistent results.

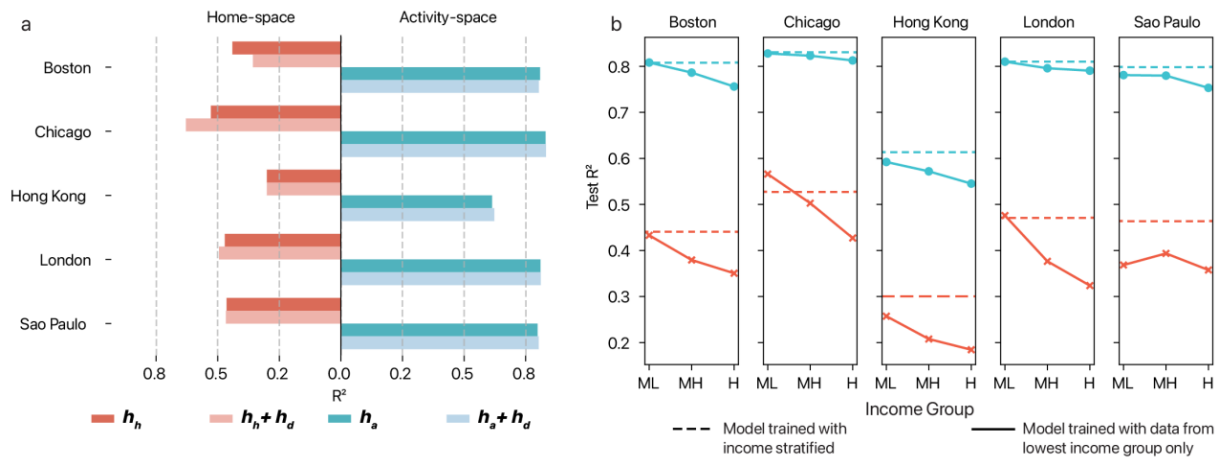

*Fig. 5 Predicting individual-level place exposure. a. The mean $R^2$ of models that include $h_h, h_h || h_d, h_a, h_a || h_d$. $||$ indicates that two embeddings are further concatenated to create a latent code for the decoder to reconstruct the place exposure. b. We use the Low-income group as the training data and train the autoencoder model and then apply the model to predict the place exposure of all other income groups. ML stands for Medium Low; MH stands for Medium High; H stands for High. The dash line indicates the model performance using random split stratified on income (75% data used for training and 25% used for testing). Full results that use other income groups as the training data and other evaluation metrics are shown in SI Fig. S23-24.*

Lastly, we retrain the model using only one income group and then test the performance on all other income groups. Suppose people's mobility pattern is independent from their socioeconomic background, we should see the model trained with one income group transferable to other income groups. On the contrary, if the structure of people's mobility pattern differs by their socioeconomic background, the model trained with one income group will not be as accurate when transferred to other income groups. Fig. 5b shows a pattern of model accuracy loss, indicating stratified visitation patterns across different income groups within the same city (Fig. 5b). Using single-income-group activity space embeddings shows 2-7% loss in accuracy, while using single-income-group home-space models leads to a drop of accuracy by 12-19%. Performance declines systematically as the income gap between the training and test groups widens (e.g., when the single-income-group for training is at the highest or lowest). Even in London and Hong Kong – where overall mixing levels appear relatively equal – we still observe performance loss when transferring visitations across groups. This indicates that while sociodemographic attributes contribute little directly to predicting mixing at the individual level, differences in where income groups travel create distinctive patterns of urban exposure, and thus unequal opportunities for social mixing (See SI Fig. S27 for model performance using other income groups as training data). Such results further emphasize the value of embedding rich socioeconomic data of mobility patterns into urban mixing or segregation studies.

Fan, Z., Loo, B.P.Y., Duarte, F., Ratti, C., & Moro, E. (2026). Latent patterns of urban mixing in mobility analysis across five global cities. *Nature Cities*, accepted.

# Discussion

By combining the socioeconomic depth of travel surveys with a cross-continent comparative analysis across five global cities (covering more than 200,000 travel diaries), our study offers several contributions. First, we validate the use of travel survey data for capturing experienced social mixing. Comparing our travel-survey-based index with one derived from home-census median household income, we find that the survey-based measure is on average 16% higher for the former. This suggests that previous reliance on combining mobility traces from mobile phones and home-location proxies could yield systematic lower estimate the of mixing people experience throughout daily travel.

Second, the rich socioeconomic information of travel surveys allows us to uncover life-course and gendered dynamics often invisible in mobility-only studies. We find support for the "second youth" or "third age" hypothesis: individuals in post-working life are as socially mixed—or more so—than those in late working life, especially after employment status is accounted for. Regarding gender, we find that while gender alone is not predictive of social mixing, individuals identified as likely caregivers (based on household composition) exhibit reduced social mixing compared to non-caregivers. This trend is particularly pronounced for women, suggesting a potential intersection of mobility constraints, household responsibilities, and career development, although future work using direct caregiving data is required to confirm any causal link.

Third, leveraging travel survey data, we identify the central role of mobility patterns and destination diversity. Place exposure at destinations explains most of the variations in social mixing, overshadowing sociodemographics, home environment, and transit proximity. Public transit use is consistently associated with higher social mixing, but this effect weakens once place exposure is controlled for, especially in Boston, Chicago, and São Paulo—implying that transit contributes to mixing primarily by connecting people to POI-rich areas. In contrast, this mediating effect is less pronounced in Hong Kong and London, where diverse amenities are more evenly distributed. Moreover, proximity to major transit stations reduces the influence of individual socioeconomic status on social mixing. However, we interpret these findings with caution, as individuals living near transit may differ systematically from those who do not, leading to a phenomenon called self-selection[36]. Nevertheless, these patterns clearly highlight the importance of urban accessibility and destination diversity in shaping opportunities for social exposure.

Furthermore, our single-income-group transferability tests via the autoencoder model reveal a critical nuance. Even though sociodemographic attributes contribute little directly to predicting individual mixing, the visitation patterns of different income groups remain stratified. These structural differences in travel patterns across income groups remain evident even in cities like London and Hong Kong, where the overall mixing levels appear rather balanced across social groups. In short, people from different income groups may share a similar level of social mixing. Yet, they experience parallel rather than shared urban experiences as people of different income groups systematically visit different places. Recent studies show that embedding activities directly into mobility models can boost both interpretability and out-of-distribution performance[37,38]. Our study joins this group of research and attempts to lay the ground for an "Urban AI" approach where theories of human behavior and mobility inform algorithmic models, improving their generality and fairness.

Finally, our cross-city comparison underscores the relationship between transit systems and social segregation. By pulling five distinctly different cities in this study, we observe that the gap in social mixing between low- and high-income is smaller in Hong Kong and London than in the other three cities. This is likely due to the rather tightly connected amenities and more homogeneous public transit usage and accessibility in these two cities so that people of different income groups have higher chances to collocate in transit hubs and public spaces. In the case of Boston and Chicago, public transit stations are located closer to lower income groups who also have lower car ownership, while in São Paulo, the highest income group has both higher access to train stations and higher car ownership, exacerbating the unequal distribution of transportation opportunities.

It is important to acknowledge the limitations of this study. First, while survey weights ensure resident-level representativeness, they may not strictly capture trip-level exposure precision. We address this via robustness checks (SI Table S2), confirming that our core findings hold across time-weighted and unweighted specifications. Additionally, we acknowledge that travel survey data, like mobile phone or other data, are not error-free. In particular, non-response and underreporting, though partially overcome by professional data weighing and expansion statistically, should be recognized[26-30]. Combining travel surveys with other big data (notably mobile phone traces) (e.g., through simulated or synthetic datasets) is a promising direction[26]. Secondly, this study focuses on social mixing only as social interaction is difficult, if not impossible, to capture without detailed observations or recordings (e.g., that a conversation has taken place or people were shaking hands)[32,33]. Future research that distinguishes trip purposes may be a step closer towards estimating potential interaction. Thirdly, travel surveys, by their nature, are cross-sectional snapshots collected at intervals (typically every decade) and often restricted to weekday activities. Thus, the social mixing patterns we capture may not reflect weekend or seasonal dynamics, not to mention dynamic changes in behavior (for example, shifts due to the COVID-19 pandemic or emerging mobility services like ride-hailing and bike-sharing). Given that the survey collection years of this study vary from 2011 to 2018 (encompassing different economic cycles and phases of transit infrastructure development), observed disparities may involve temporal context impacts that are worth further studies to confirm. Lastly, while our five-city comparison spans a diversity of global contexts, it by no means covers the full spectrum of urban environment. Replicating our analysis in other cities, especially in the developing world would further test the generalizability of our conclusions.

# Methods

## Datasets

### Five study area

The definition of city in this study is intended to be devoid of additional administrative boundaries, including urban cores or other subdivisions. For Boston, MA and Chicago, IL, the definition corresponds to the metropolitan statistical areas (MSA) within the respective states (Massachusetts and Illinois, respectively); for London, this refers to the Greater London area; for São Paulo, we use the metropolitan area of São Paulo; for Hong Kong, given its peninsula nature, we use the entire administrative region. Maps of all cities are shown in the SI Fig. S1 and S2.

### Travel surveys from five cities

Travel surveys in the selected cities, although in different formats, contain common information including person unique identifier, household unique identifier, home location (at survey zone-level), household income (in local currency), age, sex, employment status, trip purpose, trip starting time and end time, distance, trip mode (and trip leg mode), participants home location (at a street block, zone, or census block level), and expanded trip factors (See SI Section 1 for manual standardization of the employment status, trip purpose, etc.). Each participant reported multiple origins and destinations on a weekday. Each metropolitan area has different trip measures. The details of each travel survey are explained in Table 2. The finest details of travel diary are reported using different terms. To illustrate, London's survey uses "stage" to document each stop within one trip. Hong Kong's survey is leg-based – each trip can contain multiple legs. To standardize the calculation, we use trip legs as the finest unit. Table S1 includes details in original survey methods from each city.

Since the participants in these travel surveys were primarily random samples of the total population, each survey team provides an expansion factor that allow us to correct sampling bias. Previous research has demonstrated that the five surveys used in this study is well-balanced among income groups, gender and geography. We repeat our core analysis by using weighted and unweighted trips to test for the results' robustness (See SI Table S1).

From all cities, the raw survey data contains a total of 302,796 participants, among which 209,817 reported at least one trip. We applied filters to improve the reliability of the data. We removed the people who didn't have complete data regarding the home zone, income level, age, and gender reported. People who didn't report any travel were also excluded from the main analysis. All travel surveys used in this study have been used for research in the transport literature[12,39-42]. See how we standardize work status, trip purpose, trip mode in SI Section 1.

To validate the travel-survey samples, we paired each study area with the smallest readily available census-style geography that reports population and income: U.S. census tracts, U.K. Middle Super Output Areas (MSOAs), Hong Kong Tertiary Planning Units (TPUs), and São Paulo research zones. This process yields 1471 tracts in Boston, 2014 in Chicago, 1141 MSOAs in London, 289 TPUs in Hong Kong, and 517 research zones in São Paulo, with median unit areas ranging from 2.1 km² (Hong Kong) to 7 km² (Boston) (See SI Section 2.1 for details of Census survey geometry from each study city).

## Train and bus stations from OpenStreetMap

We obtain train and bus stations data from OpenStreetMap using the osmnx python package[43] by specifying the tags including bus stops, stations, and railways. A total of 82,362 bus stations and a total of 1,624 train stations are downloaded. The data was accessed in March 2024. To make sure that the station data is aligned with the travel survey data, we obtain the train stations' open date by looking up the train station names through local transportation department websites, Wikipedia, and online newspapers via ChatGPT API (model 4o). Among the 1,624 train stations, we found 63 train stations that were built during or after the travel surveys were conducted. These train stations have been removed from the household transportation opportunities measure. We cannot further identify the exact open year of the bus stops, so all bus stops are kept for this study. We acknowledge that this is a limitation of the data.

## Other POI amenities

We use POI data in each city to analyze people's exposure to different places. The primary analysis uses the POI data downloaded from OpenStreetMap. We need to standardize the raw POI tags (2,688 unique tags in total) into standard POI categories. To do so, we first use ChatGPT's 4o API to read all the POI tags and prompt to label them as one of the categories we specified: accommodation, agriculture, art & museum, café, community, education, construction, entertainment, finance, food, government, grocery, health, industry, library, office, outdoor, service, shopping, sports, tourism, transportation, utilities, worship and other. Then we manually review the categories generated from the API call. The category selection follows the previous literature[44]. For the analysis, we only include 18 categories that are relevant in our studies. Accommodation, agriculture, construction, industry, others and utilities categories are removed. We tested that our results are not significantly impacted by adding or removing one or more categories. For the robustness of the study, we also collect and analyze POI data from Safegraph (Boston and Chicago), Geo Data Store (Hong Kong), Digimap (London), Google Place API (São Paulo) to repeat the major portion of the study. See the details of the POI data verification in SI Section 3.4.

## Data Processing

### Identify gender, age, and caregiving responsibilities from travel surveys

With each participant's household member surveyed, we can identify participants that are likely to be parents or caretakers by looking at the age distribution of household members. For participants older than 21 and with at least one minor (age below 18) who are at least 18 years younger than the participants, we label them as caretakers (parents). Among all participants that are between 21 and 65 years old (195,710), 35.6% of participants are likely to be parents (38.1% after weighting by the expansion factor). Specifically, after counting the number of adults in each household, we identify 5.9% participants (9.6% after weighting) as single parents. 66% of parents (67.8% after weighting) reported travelled at least once on the survey day.

### Standardize the study unit

Travel survey data provides the origins and destinations of each trip leg. These locations are coded with local spatial units (census blocks, street blocks, latitude and longitude). To standardize the calculation, we use the H3 hexagons to code all origins and destinations for the analysis. The primary measure uses the Level 8 (average edge length of 0.53 km) H3 Hexagons for analysis. We use level 8 considering its size is comparable to census tract in the U.S. We also repeat the social mixing measure by change the H3 unit to 4, 5, 6, 7, 9 and 10 for understanding the impact of the unit resolution. We repeat the social mixing measure estimations using the census geometry from each survey area. When the hexagon size goes larger, overall social mixing level will increase for each person given they are defined to be exposed to more people within the same hexagon (see SI Fig. S6-7 for systematic comparisons of social mixing by changing the destination resolution). SI Fig. S18 shows that our results are robust against the choice of destination definition.

Fan, Z., Loo, B.P.Y., Duarte, F., Ratti, C., & Moro, E. (2026). Latent patterns of urban mixing in mobility analysis across five global cities. *Nature Cities*, accepted.

## Residential zones

The original travel surveys already provided a high-resolution residential location for each survey respondent. Their original residential zones are listed as following. Boston: census block; Chicago: census tract; Hong Kong: small street block; London: Northing and Easting imputed from the Travel Survey Unit; São Paulo: research zone. Similar to the activity locations, we transform the residential zone to H3 hexagons (level 8) and census zones with similar sizes across the cities. The transformed census zones used are the same as specified in the SI Section 2.1 section.

## Income categories

To calculate the social mixing index, we first need to label the participants with different income categories based on the rank of their household income within the selected study area (See SI Fig. S3-4 for distribution of income groups across five region). Altogether, 299,955 participants (99.06%) have available household income provided in the original travel surveys. Then, after applying expansion factors, we divide the participants into four income groups: low, mid-low, mid-high and high income. The distribution of income groups within each study area is relatively equal. To test the impact of income group definition to the social mixing index measure, we also repeat the analysis by putting participants into five income groups. In addition, to compare with the method of using home zone census income directly, we also extract each participants' home census area's median household income and repeat the analysis.

## Household transport opportunities

We calculate the walking distance of each household's home location's zone centroid to their nearest train station and bus station (See Train and Bus Stations from OpenStreetMap section for data sources). Households within 1 km of train stations or 800 meters of bus stations are considered accessible to public transit. The walking distance used here is the network distance computed using Open Source Routing Machine[44] (OSRM)'s routing API service. For each household, we also obtain self-reported information about vehicle ownership from the travel surveys.

## Major transit hubs

Our method of transit hubs identification follows the model in Roth et al.[34]. We first compute the total weighted arrivals and departures of all train stations located hexagons during morning hours (6 am − 11 am). Then we gather these spatial units by the descending order of morning arrivals. By using the hexagon unit directly, we aggregate the total arrivals that could be generated from more than one very close stations. For the mono-centric model, we only pick a single hexagon that has the highest number of arrivals as the main transit hub. For the poly-centric model, we select the top N stations that accumulate at least 60 percent of the total arrivals (See SI Section 3.1 and Fig. S13 for more details).

## Place exposure

The POI data contains place types, latitude, longitude, place names, and place unique identifier. To analyze people's exposure to different places, we create two vector representations $P_{di}$ and $P_{hi}$, to represent the list of places' counts by POI categories ($N$ = 18 categories included in our study) that are within $T$ minute walking distance of an individual $i$'s destination and home. To get all places within any location's $T$ minute walking distance, we use python *omnx* package[45] that leverages OpenStreetMap's road network to calculate the walking isochrone from each trip leg destination, and then query all places within the isochrone. For a person $i$, visiting a list of destination $D_i = \{d_1, d_2, d_3, ..., d_j\}$, the $P_{di}$ reflect the average place counts by POI categories: $P_{di} = \frac{1}{D}\sum_{d_{ij}}^{D_i} POI_j$, $POI \in R^N$. Here $POI_j$ is a vector of integers that count the number of POI by category in destination $j$ within walking distance $T$. The main analysis uses $T$ = 15 minute to report the results. A visualization of the POI query can be found in SI Fig. S12. We also test $T$ = 5 or 10 and the results are included in the SI Fig. S20b and c. In addition, we also repeat the analysis by completely replacing the POI sources with POI data source from each local POI provider (SI Fig. S18).

## Predicting place exposure

### Model details

The aim of this analysis is to predict individual daily exposure to different categories of urban places, represented as an $N$-dimensional vector where each dimension corresponds to the count of potential visits to a specific place category ($N$ being the total number of place categories included in this study). We do not know exactly which POI (e.g., the name of the shop or a facility) the person visits at destinations. Therefore, we query all POIs within 15-min walking distance of each trip destination to represent the places that they are exposed to. We also tested 5-min and 10-min walking distance for robustness.

To predict the daily exposure, we integrate the information about individuals' socioeconomic factors, home locations, daily travel activity areas, and the spatial and transit characteristics of each city. Our model leverages three key insights about urban networks and human mobility: (1) places that are geographically closer influence each other more significantly[46,47]; 2) places can also have important long-distance connections facilitated by human mobility[44]; 3) public transit accessibility significantly impacts people's place exposure[20]. To represent these relationships, we first construct a spatio-temporal place network for each city, incorporating both physical proximity based on road networks and travel time derived from public transit schedules (GTFS data). We embed the detailed network information into each place using a graph neural network (GNN)[48] (Fig. 4a). A total of 107,750 POIs with 12.7 million edges are created. Next, we employ a supervised autoencoder[49] neural network designed in an encoder-decoder structure (Fig. 4b and c). Unlike traditional autoencoders, which compress and reconstruct their input exactly, our autoencoder aims to learn a concise, meaningful representation of input features to predict the individual place exposure vector. Specifically, the encoder processes individuals' sociodemographic attributes and their associated place network features (either around home or throughout their daily activity space), as mapped from their travel trajectories to create a compact latent embedding. The decoder then uses this embedding to output the predicted place exposure vector, quantifying the expected number of urban amenities a person is exposed to (see Fig. 4).

## Construct POI place embeddings with GNN

To represent each city's spatial network information with POI places, we first build a POI graph $G(v, e, w)$ for each city that uses each POI as the node $v$, the edge $w_{ij}$ measures the connectivity between two nodes $v_i$ and $v_j$. For the robustness of the study, we build four different graph $G$ using road network attributes and transit availability (using public GTFS data). The primary measure uses the transit-time-aware network that finds $v_j$ within 15-minute walking or transit time of $v_i$. We assign $w_{ij} = 1$ if within the 15-minute travel time, 0 otherwise. On average, this network contains 8.7 million to 39.7 million edges in all cities. The average edge per node ranges from 33.6 (Boston) to 230.9 (London). The second method creates network edges that are time-weighted ($w_{ij} = e^{-t_{ij}/\alpha}$), where $t_{ij}$ is the travel time between nodes $v_i$ and $v_j$. In addition, we also construct two more distance-based measures. The third one uses $w_{ij} = 1$ to indicate that within 1.26 km (approximately 15-minute walking distance) of node $v_i$, we can find $v_j$. The fourth measure constructs the $w_{ij} = e^{-d_{ij}/\alpha}$ to further indicate the strength of spatial proximity, where $d_{ij}$ is the distance between the two POI places, $\alpha$ here is a scale factor. The GNN model performance by four edge types is shown in SI Table S7. Our results are robust against variations on spatial graph construction (See SI Section 5.1 for detail construction of the model). With the network established, we train a two-layer Graph Coevolutionary Neural Network model (one type of GNN model) to predict the category of each POI place (See SI Section 6.1). The contribution of GNN embedding is shown in Table S11.

It is important to clarify that the target variable of daily exposure and the input variable of place embedding in the GNN model capture fundamentally different geometric and statistical dimensions of the spatial data. The target variable is an aggregate categorical frequency distribution representing the density and diversity of potential encounters with people of the same and other income groups specifically at discrete destination points where an individual stopped. In contrast, the activity-space embedding ($h_a$) generated by the Graph Neural Network encodes the latent spatial topology of the wider geographical contexts within all potentially accessible areas of an individual. Because $h_a$ is constructed from all POIs within the spatial convex hull encompassing people's full daily trajectories, it represents the broader geographic boundaries of their movements, rather than just their chosen stops. Consequently, the autoencoder does not perform a mechanical aggregation. Rather, it offers an interpretable test on a structural hypothesis: whether the topological shape and total area of a person's movement network intrinsically determines their specific categorical exposure profile. The non-trivial nature of this distinction is evidenced by our transferability tests, which show that the relationship between place exposure and spatial territory is highly complex and stratified by income.

## Data Availability

The aggregated data for repeating the analysis will be shared via this repository (https://github.com/brookefzy/social-mixing-5-city) upon paper acceptance.

## Tables

*Table 1 Estimate Daytime Social Mixing with Transit Modes*

| | DAYTIME SOCIAL MIXING (DM) | | | | | | | | | |
|---|---|---|---|---|---|---|---|---|---|---|
| **VARIABLE** | Boston | | Chicago | | Hong Kong | | London | | São Paulo | |
| | Model 1 | Model 2 | Model 1 | Model 2 | Model 1 | Model 2 | Model 1 | Model 2 | Model 1 | Model 2 |
| **% PUBLIC TRANSIT** | 0.337*** | 0.091*** | 0.142*** | 0.044*** | 0.140*** | 0.127*** | 0.167*** | 0.128*** | 0.107*** | 0.060*** |
| | (0.016) | (0.013) | (0.012) | (0.011) | (0.005) | (0.005) | (0.004) | (0.004) | (0.002) | (0.002) |
| **% PRIVATE TRANSIT** | -0.022*** | 0.027*** | 0.031*** | 0.064*** | 0.076*** | 0.075*** | -0.066*** | -0.049*** | 0.008*** | -0.003*** |
| | (0.007) | (0.006) | (0.000) | (0.000) | (0.000) | (0.000) | (0.000) | (0.000) | (0.000) | (0.000) |
| **OBS** | 7623 | 7623 | 18257 | 18257 | 42736 | 42736 | 30382 | 30382 | 46957 | 46957 |
| **R²** | 0.3448 | 0.5677 | 0.1077 | 0.3001 | 0.0724 | 0.2877 | 0.1428 | 0.2208 | 0.1122 | 0.2899 |
| **Control S+H** | Yes | Yes | Yes | Yes | Yes | Yes | Yes | Yes | Yes | Yes |
| **Control PD+PH** | No | Yes | No | Yes | No | Yes | No | Yes | No | Yes |

*Notes: This table reports results of two variations of Equation 3. The Model 1 estimates daytime social mixing with S+H+M. Model 2 estimates daytime mixing with S+H+M+PD+HD. Robust standard error in parenthesis. \*\*\* represents p-value <0.001.*

*Table 2 Travel Survey Summary*

| Study Region | Boston, MA, US | Chicago, IL, US | Hong Kong, SAR | London, UK | São Paulo, Brazil |
|---|---|---|---|---|---|

Fan, Z., Loo, B.P.Y., Duarte, F., Ratti, C., & Moro, E. (2026). Latent patterns of urban mixing in mobility analysis across five global cities. *Nature Cities*, accepted.

| Site Description | Boston Metropolitan area within MA | Chicago Metropolitan Area within IL | Hong Kong Special Administrative Region (HKSAR) | Greater London area/32 London boroughs and the City of London | Metropolitan area of São Paulo |
|---|---|---|---|---|---|
| Year of Survey | 2009-2011 | 2017-2018 | 2010-2011 | 2012-2014 | 2017 |
| Survey Duration | 17 months | 4 months | 4.5 months | continuous rolling survey | 16 months |
| Travel Survey Name | Massachusetts Travel Survey | My Daily Travel | Travel Characteristics Survey | London Travel Diary (LTD) | São Paulo Travel Diary |
| Original Survey Trip Reported Resolution | Census Block | Census Tract | Small Street Block | Northing, Easting | Coordinates |
| Travel Survey Methods | - Computer-Assisted Telephone Interviewing (CATI) for recruitment and retrieval<br>- Mail-back diaries option (52.5% mailed back, 47.5% phone retrieval)<br>- Advance letter followed by telephone contact | - Primarily web-based recruitment surveys<br>- Smartphone app option (Daily Travel App) for recording travel<br>- Mail invitations with online completion | - Self-administered questionnaire with option for face-to-face interview upon request<br>- Hotline available for assistance | - Face-to-face household interviews<br>- Three questionnaires used: household questionnaire, individual questionnaires, and trip sheets/travel diaries<br>- All completed in-person | - Two-part survey: Household Survey (domestic) + Boundary Line Survey (external trips)<br>- Tablet-based application for interviews<br>- Multiple interview locations (roadways, bus terminals, airports, metro stations) |
| Survey Resolution | Trip | Stop | Trip Leg | Stage | Stop |
| Population of the year (Million) | 4.24 | 5.17 | 6.88 | 8.6 | 21 |
| Total households included | 10,785 | 12,391 | 35,401 | 24,248 | 31,847 |
| Total participants interviewed | 26,770 | 30,683 | 101,385 | 57,640 | 86,318 |
| Total weighted participants (M) | 4.03 | 6.67 | 6.31 | 8.95 | 16.7 |

*Notes: This table reports the detail of each travel survey data. London and São Paulo's survey data contains coordinates-level information for each trip thus we don't provide the original resolution size here. Expanded table including details in data administration from each original survey is shown in Table S1.*

Fan, Z., Loo, B.P.Y., Duarte, F., Ratti, C., & Moro, E. (2026). Latent patterns of urban mixing in mobility analysis across five global cities. *Nature Cities*, accepted.

# Extended data figures

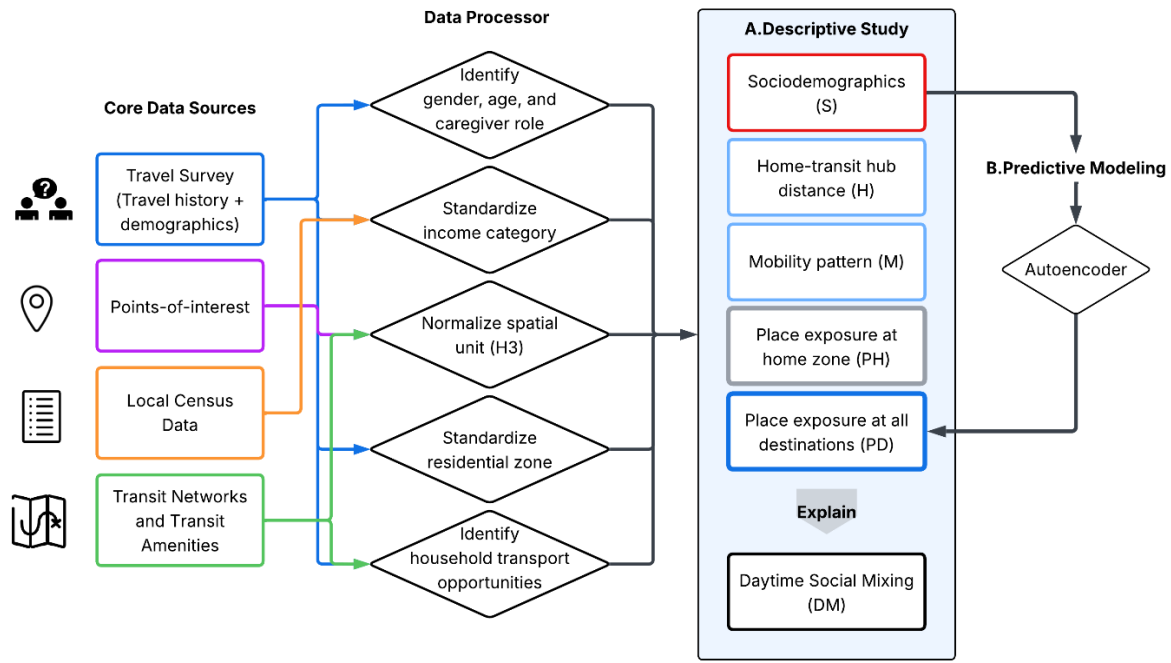

*Fig. E 1 Schematic data processing pipeline. Details in how the transportation modes, work status, and trip purposes are standardized are included in SI Table S8-10.*

# References

1. Ratti, C., Frenchman, D., Pulselli, R. M. & Williams, S. Mobile Landscapes: Using Location Data from Cell Phones for Urban Analysis. *Environ. Plan. B Plan. Des.* **33**, 727–748 (2006).

2. Eagle, N. & Pentland, A. S. Eigenbehaviors: identifying structure in routine. *Behav. Ecol. Sociobiol.* **63**, 1057–1066 (2009).

3. Moro, E., Calacci, D., Dong, X. & Pentland, A. Mobility patterns are associated with experienced income segregation in large US cities. *Nat. Commun.* **12**, 4633 (2021).

4. Wang, A. et al. Personal Mobility Choices and Disparities in Carbon Emissions. *Environ. Sci. Technol.* **57**, 8548–8558 (2023).

5. Zhang, Y., Fu, W., Chao, H., Mi, Z. & Kong, H. A comparative analysis of the potential of carbon emission reductions from shared micro-mobility. *Sustain. Energy Technol. Assess.* **72**, 104088 (2024).

Fan, Z., Loo, B.P.Y., Duarte, F., Ratti, C., & Moro, E. (2026). Latent patterns of urban mixing in mobility analysis across five global cities. *Nature Cities*, accepted.

6. Nyhan, M. M., Kloog, I., Britter, R., Ratti, C. & Koutrakis, P. Quantifying population exposure to air pollution using individual mobility patterns inferred from mobile phone data. *J. Expo. Sci. Environ. Epidemiol.* **29**, 238–247 (2019).

7. Loo, B. P. Y. & Axhausen, K. Getting out of energy-intensive and "dirty" transport. *Innovation* **3**, 100339 (2022).

8. Ewing, R., Tian, G. & Lyons, T. Does compact development increase or reduce traffic congestion? *Cities* **72**, 94–101 (2018).

9. Sardari, R., Hamidi, S. & Pouladi, R. Effects of Traffic Congestion on Vehicle Miles Traveled. *Transp. Res. Rec.* **2672**, 92–102 (2018).

10. Ma, J., Rao, J., Kwan, M.-P. & Chai, Y. Examining the effects of mobility-based air and noise pollution on activity satisfaction. *Transp. Res. Part D Transp. Environ.* **89**, 102633 (2020).

11. Yin, X., Fallah-Shorshani, M., McConnell, R., Fruin, S. & Franklin, M. Predicting Fine Spatial Scale Traffic Noise Using Mobile Measurements and Machine Learning. *Environ. Sci. Technol.* **54**, 12860–12869 (2020).

12. Lian, T., Loo, B. P. Y. & Fan, Z. Advances in estimating pedestrian measures through artificial intelligence: From data sources, computer vision, video analytics to the prediction of crash frequency. *Comput. Environ. Urban Syst.* **107**, 102057 (2024).

13. Loo, B. P. Y., Tsoi, K. H., Wong, P. P. & Lai, P. C. Identification of superspreading environment under COVID-19 through human mobility data. *Sci. Rep.* **11**, 4699 (2021).

14. Florida, R. The flight of the creative class: The new global competition for talent. *Lib. Educ.* **92**, 22–29 (2006).

15. Major, B., Mendes, W. B. & Dovidio, J. F. Intergroup relations and health disparities: a social psychological perspective. *Health Psychol.* **32**, 514–524 (2013).

16. Chetty, R. et al. Social capital I: measurement and associations with economic mobility. *Nature* **608**, 108–121 (2022).

17. Cabrera, J. F. & Najarian, J. C. How the Built Environment Shapes Spatial Bridging Ties and Social Capital. *Environ. Behav.* **47**, 239–267 (2015).

18. United Nations. *SDG Indicators*. https://unstats.un.org/sdgs/indicators/indicators-list/ (2023).

19. Nilforoshan, H. et al. Human mobility networks reveal increased segregation in large cities. *Nature* **624**, 586–592 (2023).

Fan, Z., Loo, B.P.Y., Duarte, F., Ratti, C., & Moro, E. (2026). Latent patterns of urban mixing in mobility analysis across five global cities. *Nature Cities*, accepted.

20. Loo, B. P. Y., Fan, Z. & Moro, E. Residential and experienced social segregation: the roles of different transport modes, metro extensions, and longitudinal changes in Hong Kong. *Humanit. Soc. Sci. Commun.* **11**, 1439 (2024).

21. Athey, S., Ferguson, B., Gentzkow, M. & Schmidt, T. Estimating experienced racial segregation in US cities using large-scale GPS data. *Proc. Natl. Acad. Sci. U. S. A.* **118**, e2026160118 (2021).

22. Han, Y., Liao, P., Li, W. & Wang, Q. R. Visitation patterns reveal service access disparities for ageing populations in the USA. *Nat. Hum. Behav.* https://doi.org/10.1038/s41562-025-02285-4 (2025).

23. Cook, C., Currier, L. & Glaeser, E. Urban mobility and the experienced isolation of students. *Nat. Cities* **1**, 73–82 (2024).

24. Jung, H., Lee, J. S. & Kim, S. Beyond Population Density: A New Framework for Compact Cities through Amenity Survival Analysis. *Cities* **168**, 105535 (2026).

25. Jenks, M. & Burgess, R. *Compact Cities: Sustainable Urban Forms for Developing Countries* (Routledge, 2000).

26. Chen, C., Bian, L. & Ma, J. From Traces to Trajectories: How Well Can We Guess Activity Locations from Mobile Phone Traces? *Transp. Res. Part C Emerg. Technol.* **46**, 326–337 (2014).

27. National Research Council Transportation Research Board. *Innovations in Travel Survey Methods*. Transportation Research Record no. 1412 (National Academy Press, 1993).

28. Armoogum, J. & Madre, J.-L. Weighting or imputations? The example of nonresponses for daily trips in French NPTS. *J. Transp. Stat.* **1**, 53–63 (1998).

29. Jones, P. & Stopher, P. R. (eds) *Transport Survey Quality and Innovation* (Pergamon, 2003).

30. Moore, J. C. & Durrant, G. *Non-response bias risks in sample sub-groups in Understanding Society: the UK Household Longitudinal Study*. *Understanding Society Working Paper Series 2025-03* https://www.understandingsociety.ac.uk/wp-content/uploads/working-papers/2025-03.pdf (2025).

31. Bousquin, J. Discrete Global Grid Systems as scalable geospatial frameworks for characterizing coastal environments. *Environ. Model. Softw.* **146**, 105210 (2021).

32. Salazar-Miranda, A. et al. Exploring the social life of urban spaces through AI. *Proc. Natl. Acad. Sci. U. S. A.* **122**, e2424662122 (2025).

33. Loo, B. P. Y. & Fan, Z. Social interaction in public space: Spatial edges, moveable furniture, and visual landmarks. *Environ. Plan. B Urban Anal. City Sci.* **50**, 2510–2526 (2023).

Fan, Z., Loo, B.P.Y., Duarte, F., Ratti, C., & Moro, E. (2026). Latent patterns of urban mixing in mobility analysis across five global cities. *Nature Cities*, accepted.

34. Roth, C., Kang, S. M., Batty, M. & Barthélemy, M. Structure of Urban Movements: Polycentric Activity and Entangled Hierarchical Flows. *PLoS ONE* **6**, e15923 (2011).

35. Lindeman, R. H. *Introduction to Bivariate and Multivariate Analysis* (Scott Foresman, 1980).

36. Aston, L. et al. Addressing Transit Mode Location Bias in Built Environment-Transit Mode Use Research. *J. Transp. Geogr.* **87**, 102819 (2020).

37. Weng, G., Kim, M., Ahn, Y.-Y. & Moro, E. *Beyond Distance: Mobility Neural Embeddings Reveal Visible and Invisible Barriers in Urban Space*. Preprint at https://doi.org/10.48550/arXiv.2506.24061 (2025).

38. Cabanas-Tirapu, O., Danús, L., Moro, E., Sales-Pardo, M. & Guimerà, R. Human mobility is well described by closed-form gravity-like models learned automatically from data. *Nat. Commun.* **16**, 1336 (2025).

39. Jiang, S., Ferreira, J. & González, M. C. Clustering daily patterns of human activities in the city. *Data Min. Knowl. Discov.* **25**, 478–510 (2012).

40. de Sá, T. H., Parra, D. C. & Monteiro, C. A. Impact of travel mode shift and trip distance on active and non-active transportation in the São Paulo Metropolitan Area in Brazil. *Prev. Med. Rep.* **2**, 183–188 (2015).

41. Fairnie, G. A., Wilby, D. J. & Saunders, L. E. Active travel in London: the role of travel survey data in describing population physical activity. *J. Transp. Health* **3**, 161–172 (2016).

42. Forrest, T. L. & Pearson, D. F. Comparison of Trip Determination Methods in Household Travel Surveys Enhanced by a Global Positioning System. *Transp. Res. Rec.* **1917**, 63–71 (2005).

43. Boeing, G. OSMnx: New methods for acquiring, constructing, analyzing, and visualizing complex street networks. *Comput. Environ. Urban Syst.* **65**, 126–139 (2017).

44. Yabe, T., García Bulle Bueno, B., Frank, M. R., Pentland, A. & Moro, E. Behaviour-based dependency networks between places shape urban economic resilience. *Nat. Hum. Behav.* **9**, 496–506 (2025).

45. Luxen, D. & Vetter, C. Real-time routing with OpenStreetMap data. in *Proceedings of the 19th ACM SIGSPATIAL International Conference on Advances in Geographic Information Systems 513–516* (ACM, 2011).

46. Goodchild, M. F. First law of geography. in *International Encyclopedia of Human Geography* (eds Kitchin, R. & Thrift, N.) 179–182 (Elsevier, 2009).

47. Tobler, W. On the First Law of Geography: A Reply. *Ann. Assoc. Am. Geogr.* **94**, 304–310 (2004).

Fan, Z., Loo, B.P.Y., Duarte, F., Ratti, C., & Moro, E. (2026). Latent patterns of urban mixing in mobility analysis across five global cities. *Nature Cities*, accepted.

48. Scarselli, F., Gori, M., Tsoi, A. C., Hagenbuchner, M. & Monfardini, G. The graph neural network model. *IEEE Trans. Neural Netw.* **20**, 61–80 (2008).

49. Zhai, J., Zhang, S., Chen, J. & He, Q. Autoencoder and its various variants. in *2018 IEEE International Conference on Systems, Man, and Cybernetics (SMC) 415–419* (IEEE, 2018).

# Correspondence

Correspondence should be addressed to Professor Becky P.Y. Loo at bpyloo@hku.hk. The aggregated data for repeating the analysis are available via this repository (https://github.com/brookefzy/social-mixing-5-city) maintained by Dr Zhuangyuan Fan.

# Acknowledgements

This was an initiative of the University of Hong Kong (HKU) Joint Lab on Future Cities. The authors are grateful for the support of the Hong Kong Postgraduate Fellowship (HKPF), which has supported the PhD study of the first author at HKU.

# Author contributions

Z. Fan contributed to the datasets, formal analysis, validation, visualization and writing – original draft, and editing and reviewing; B.P.Y. Loo contributed to the conceptualization, datasets, supervision and writing – original draft, and editing and reviewing; F. Durate and C. Ratti contributed to the datasets, supervision and writing – editing and reviewing; E. Moro contributed to the methodology, supervision and writing – editing and reviewing.